\pdfoutput=1

\documentclass[11pt]{article}

\usepackage[preprint]{acl}
 \usepackage{multirow} 
\usepackage{times}
\usepackage{latexsym}
\usepackage{booktabs}
\usepackage{fontawesome}
\usepackage{float}
\usepackage{booktabs}  
\usepackage{caption}   
\usepackage{array}
\usepackage{amsmath}
\usepackage{amssymb} 
\usepackage{algorithm}
\usepackage{algpseudocode}

\usepackage[T1]{fontenc}

\usepackage[utf8]{inputenc}

\usepackage{microtype}

\usepackage{inconsolata}
\usepackage{subcaption} 
\usepackage{graphicx}
\usepackage{makecell}
\usepackage{amsmath}   
\usepackage{algorithm}
\usepackage{algpseudocode}
\usepackage{float}
\usepackage{algorithmicx}
\usepackage{caption}
\title{A Cognitive Writing Perspective for Constrained Long-Form \\Text Generation}

\author{First Author \\
  Affiliation / Address line 1 \\
  Affiliation / Address line 2 \\
  Affiliation / Address line 3 \\
  \texttt{email@domain} \\\And
  Second Author \\
  Affiliation / Address line 1 \\
  Affiliation / Address line 2 \\
  Affiliation / Address line 3 \\
  \texttt{email@domain} \\}

\author{
 \textbf{Kaiyang Wan\textsuperscript{1}},
 \textbf{Honglin Mu\textsuperscript{1}},
 \textbf{Rui Hao\textsuperscript{2}},
 \textbf{Haoran Luo\textsuperscript{3}},
 \textbf{Tianle Gu\textsuperscript{1}}, 
 \textbf{Xiuying Chen\textsuperscript{1}\thanks{Corresponding author.}},
\\
\\
 \textsuperscript{1}MBZUAI,
 \textsuperscript{2}University of Chinese Academy of Sciences,
 \textsuperscript{3}Nanyang Technological University
\\
  \small{{\{Kaiyang.Wan, Xiuying.Chen\}@mbzuai.ac.ae}
}}

\begin{document}
\maketitle

\begin{abstract}
Like humans, Large Language Models (LLMs) struggle to generate high-quality long-form text that adheres to strict requirements in a single pass. 
This challenge is unsurprising, as successful human writing, according to the Cognitive Writing Theory, is a complex cognitive process involving iterative \textit{planning}, \textit{translating}, \textit{reviewing}, and \textit{monitoring}. 
Motivated by these cognitive principles, we aim to equip LLMs with human-like cognitive writing capabilities through CogWriter, a novel training-free framework that transforms LLM constrained long-form text generation into a systematic cognitive writing paradigm. Our framework consists of two key modules: (1) a Planning Agent that performs \textit{hierarchical planning} to decompose the task, and (2) multiple Generation Agents that execute these plans in parallel. The system maintains quality via continuous \textit{monitoring} and \textit{reviewing} mechanisms, which evaluate outputs against specified requirements and trigger necessary revisions.
CogWriter demonstrates exceptional performance on LongGenBench, a benchmark for complex constrained long-form text generation. 
Even when using Qwen-2.5-14B as its backbone, CogWriter surpasses GPT-4o by 22\% in complex instruction completion accuracy while reliably generating texts exceeding 10,000 words.
We hope this cognitive science-inspired approach provides a paradigm for LLM writing advancements: \faGithub  \href{https://github.com/KaiyangWan/CogWriter}{CogWriter}.
\end{abstract}

\section{Introduction}

LLMs like ChatGPT \cite{achiam2023gpt} have begun to mirror human-like capabilities across diverse natural language processing tasks~\cite{xi2023risepotentiallargelanguage,luo2024text2nkg}. From crafting concise summaries~\cite{chen2025evaluating,chen2024flexible} to composing structured reports~\cite{schmidgall2025agentlaboratoryusingllm,2024autosurvey}, these models can generate coherent text in a single pass~\cite{rasheed2025largelanguagemodelscode,minaee2024largelanguagemodelssurvey} with a fluency that often rivals human writers. Recent advances have led to models with expanded context windows of up to 128K tokens~\cite{pawar2024whatwhycontextlength}, theoretically enabling the generation of extensive documents~\cite{bai2024longwriterunleashing10000word}. However, these models face significant challenges when tasked with generating constrained long-form text under complex constraints, such as following detailed instructions over 10,000 words~\cite{wu2024longgenbench}. This limitation poses a crucial barrier for applications requiring extended~\cite{shi2024segmentlongtextprocessing}, well-structured content, including creative design proposals, technical documentation, and comprehensive research reports.

\begin{figure*}[t!]
    \centering
  \includegraphics[width=0.9\linewidth]{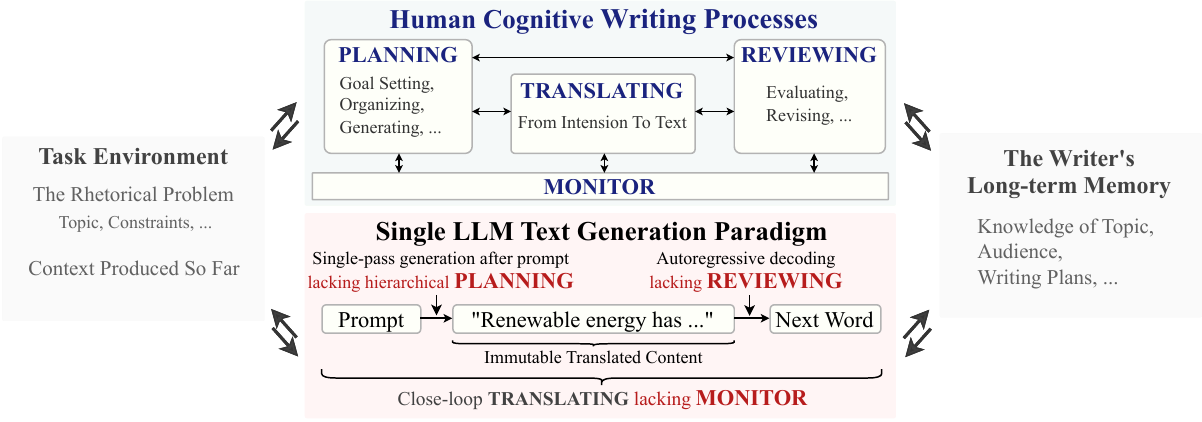} 
  \caption {Comparison of human cognitive writing processes and single LLM text generation paradigm.}
  \label{comp}
\end{figure*}

To understand the disparity between LLMs and human writers, we refer to Cognitive Writing Theory~\cite{flower1981cognitive}, which emphasizes how humans succeed in writing through a recursive activity that dynamically integrates multiple cognitive processes.
As shown in the top part of Figure~\ref{comp}, these processes include \textit{planning}, where writers establish high-level goals and develop structural outlines; \textit{translating}, where writers transform abstract ideas into coherent text; and \textit{reviewing}, where writers continuously evaluate and refine their generated content. Crucially, writers control these components through continuous \textit{monitoring}, allowing them to assess and adjust text to better align with evolving objectives throughout the writing process.

Current LLMs excel at generating fluent text, effectively performing the \textit{translating} function of converting internal token vectors into textual content. 
However, they fundamentally conflict with key cognitive principles in three ways, as shown in the bottom part of Figure~\ref{comp}: \textit{1)} They treat long-form text generation merely as an end-to-end task, overlooking the crucial hierarchical \textit{planning} process that should guide content generation; \textit{2)} Their autoregressive architecture renders generated tokens as immutable context, preventing the \textit{reviewing} and restructuring capabilities essential to human writing; and \textit{3)} Unlike human writers who actively \textit{monitor} their progress against both local and global objectives, LLMs lack explicit evaluation mechanisms, leading to potential divergence from intended goals in extended generations.

To address the limitations of single-pass generation, we introduce CogWriter, a novel training-free framework that aligns LLM-based text generation with cognitive writing paradigm. 
At its core, CogWriter employs a Planning Agent that decomposes complex requirements into manageable subtasks, providing explicit guidance for content generation. 
Based on sub-plans and the initial goal, multiple Generation Agents work in parallel to produce text segments, enabling both efficient generation and quality control that ensures consistent alignment with requirements.
Crucially, both the planning and generation processes support iterative reviewing through feedback from external monitoring functions and LLM-based evaluation, thus enabling dynamic plan adjustment and content revision. 

We evaluate CogWriter on LongGenBench-16K~\cite{wu2024longgenbench}, a benchmark designed to test a language model's ability to generate instruction-aligned content about 16K tokens. 
Empirical results demonstrate that our paradigm is effective for both closed-source and open-source LLMs of various sizes. 
Specifically, even when using Qwen-2.5-14B as its backbone, CogWriter achieves a 22\% higher instruction completion accuracy rate compared to GPT-4o, while reliably generating texts exceeding 10,000 words.
These results demonstrate the effectiveness of cognitive science-inspired approaches in advancing LLM writing capabilities, particularly for complex constrained long-form text generation. 
We hope CogWriter's systematic cognitive writing paradigm will inspire future research in LLM writing advancement.

Our contributions can be summarized as follows:  
\begin{itemize}
    \item We provide a cognitive science perspective on the shortcomings of single-pass LLM generation, highlighting how it diverges from established successful human writing processes.
    \item We propose CogWriter, a cognitive writing framework that equips LLMs with human writing strategies using multiple LLM-based agents with external monitoring functions.
    \item We demonstrate that CogWriter remarkably enhances LLMs' ability to produce long-form, instruction-compliant texts without requiring additional training or reinforcement learning.
\end{itemize}

\section{A Cognitive Writing Perspective}
\label{analysis}
The challenge of constrained long-form text generation extends far beyond simply producing more words. 
Just as a novelist crafts an intricate narrative or an architect designs a towering structure, long text generation requires the coordination of multiple cognitive processes working together.
Through the lens of cognitive writing theory, three fundamental processes emerge: hierarchical planning, continuous monitoring, and dynamic reviewing~\cite{flower1981cognitive}, as illustrated in Figure~\ref{comp}.

\paragraph{Hierarchical Planning}
Long-form writing requires a delicate cognitive balance between maintaining local coherence and global structure. Human writers cope with this constraint, as working memory cannot simultaneously retain every detail of a complex narrative~\cite{kellogg2013model}.
Skilled writers manage this limitation through hierarchical decomposition, systematically structuring the writing process into multiple levels (e.g., chapters, sections, and paragraphs). This approach enables them to alternate between top-down thematic planning and bottom-up content development, ensuring alignment with high-level objectives while refining details~\cite{hayes2016identifying}.

LLMs encounter a similar limitation: they generate text in a linear, autoregressive manner without an independent planning module to iteratively refine outlines or adapt strategies in real time~\cite{xie-etal-2023-next}. Consequently, their direct prompt-to-text generation process often struggles with complex, multi-threaded narratives. Without structured guidance, LLMs are prone to losing coherence over long spans, as their finite computational capacity quickly becomes overwhelmed~\cite{hu2024hiagenthierarchicalworkingmemory}.

\paragraph{Continuous Monitoring}
Effective planning in writing requires continuous oversight. Human writers naturally monitor their work, acting like their own editors. They pay attention to both small details—such as word choice and sentence flow—and the larger structure, ensuring the text maintains a clear theme and purpose~\cite{kellogg2013model}.

In contrast, current mainstream LLMs generate text in a linear, close-loop manner, without the ability to review or refine their output. They lack a built-in system to check their progress against the intended goals, making it difficult to spot and correct issues during generation. Without external monitoring, LLMs struggle to detect when the content drifts off-topic, when the style becomes inconsistent, or when repetition occurs—problems that are especially common in extended long-form writing~\cite{wang2024generatinglongformstoryusing, ping2025longdpounlockbetterlongform}.

\paragraph{Dynamic Reviewing}
While monitoring continuously tracks the writing process by detecting small errors, inconsistencies, or deviations, reviewing takes this feedback and applies it to make necessary adjustments, such as reorganizing content or improving logical flow.
Human writers naturally engage in this iterative reviewing process, refining their work by revisiting earlier content and making adjustments~\cite{bereiter2013psychology}.

However, LLMs lack this ability due to their left-to-right, single-pass generation~\cite{10.5555/3666122.3666639,wu-etal-2024-large}. 
Without the ability to revisit or reorganize previous content, LLMs struggle with global revisions, such as restructuring sections or ensuring consistency across distant parts of the text~\cite{bae-kim-2024-collective,cheng2024sparselfplaytreesearchrefinement,cheng2024lift}. This absence of dynamic reviewing often results in long-form outputs with accumulated errors, inconsistencies, or redundant content.

\begin{figure*}[t!]
\includegraphics[width=1\linewidth]{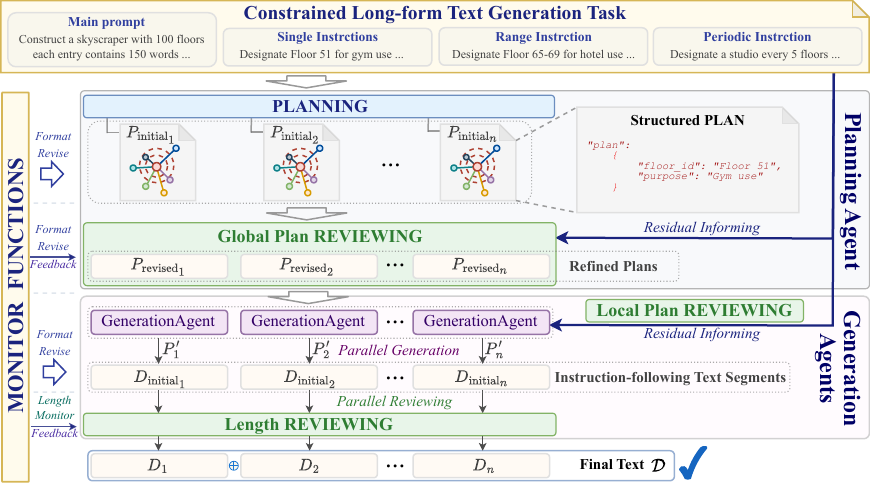}
\caption{\textbf{Overview of the CogWriter Framework.} The framework consists of two key modules: the Planning Agent and the Generation Agents.
The Planning Agent generates and refines an initial plan, guiding the structure and flow of the document.
The Generation Agents collaborate to generate, revise, and finalize document segments, ensuring consistency in content and narrative coherence across the entire document.}
\label{framework}
\end{figure*}

\section{Problem Formulation}
\label{probdef}

Based on the analysis in Section~\ref{analysis}, successfully generating long-form text requires addressing key deficiencies in current LLMs.  
We propose a new paradigm that equips LLMs with essential abilities to handle long, complex, and instruction-driven text generation.  
To achieve this, we formally define the constrained long-form text generation task, specifying the types of instructions and requirements the model must meet.

Following \citet{wu2024longgenbench}, we formally define constrained long-form generation as the task of generating a sequence of interrelated text segments $\mathcal{D} = \{D_1, D_2, \dots, D_n\}$, where each $D_i$ represents a coherent unit of text that must satisfy certain constraints. Each segment $D_i$ must achieve a target $L$ words and adhere to a set of instructions $\mathcal{T}$.
The instructions $\mathcal{T}$ guide the generation process and are classified into three types:
1. \textit{\textbf{Single Instruction (SI)}}: This instruction specifies content that must appear at exact, predefined positions. It is denoted as \( \mathcal{T}_S = \{T_{s1}, T_{s2}, \ldots\} \), where each \( T_{si} \) indicates specific content that must be placed in a precise position within the generated descriptions.  
2. \textit{\textbf{Range Instruction (RI)}}: This instruction specifies the content that must be included in each description within a designated range. It is represented as \( \mathcal{T}_R = \{T_i, T_{i+1}, \ldots, T_{i+j}\} \), ensuring that the specified content is sequentially assigned within the range \( [i, i+j] \).  
3. \textit{\textbf{Periodic Instruction (PI)}}: This instruction mandates the periodic repetition of specific content at regular intervals. It is defined as \( \mathcal{T}_P = \{T_n, T_{2n}, \ldots, T_{m \cdot n}\} \), where \( n \) is the interval length and \( m \) specifies the number of repetitions.
These instructions are unified into a comprehensive Check Set: $\mathcal{T} = \{\mathcal{T}_S, \mathcal{T}_R, \mathcal{T}_P\}$.

The versatility of this framework extends to various practical applications. For example, in architectural planning for a 100-floor building, Single Instructions determine specific facilities like a medical center on the 20th floor, Range Instructions define functional zones like corporate offices spanning floors 5-12, and Periodic Instructions maintain consistent amenities such as security checkpoints on every fifth floor. Each floor description must meet a target length of 200 words.

\section{Methodology}

Drawing upon our analysis of cognitive writing processes and the identified limitations of single-pass generation approaches, in this section, we propose CogWriter, a training-free framework that equips LLM with cognitive writing capabilities and enables LLMs to tackle complex constrained long-form generation with human-like strategic thinking.

\subsection{Framework Overview}

As shown in Figure~\ref{framework}, CogWriter is designed to bridge the gap between current LLMs and human-like writing processes by integrating planning, monitoring, and reviewing mechanisms into the generation workflow.
At its core, CogWriter employs a specialized Planning Agent that hierarchically decomposes the task and create structured plans, breaking down complex writing tasks into manageable components while maintaining their intricate relationships.
Generation Agents execute these plans while monitoring mechanisms continuously evaluate the output to detect deviations in content, structure, or requirements. 
When issues are identified by monitor or LLM, a review process is triggered to revise and refine the output, ensuring overall coherence and adherence to instructions.


\subsection{Planning Agent}

The Planning Agent serves as the strategic brain of the system. 
Similar to how an experienced writer begins with a detailed outline, this agent analyzes task requirements and generates a structured initial plan \( \mathcal{P}_\text{initial} \) under strict format constraints:  
\[
    \mathcal{P}_\text{initial} \gets \text{GenerateInitialPlan}(p_{\text{plan}}),
\]  
where \( p_{plan} \) is the task-specific prompt incorporating instruction descriptions \( \mathcal{T} \). The target plan is hierarchical, comprising unit plans: \( \mathcal{P}_\text{initial} = \{P_{\text{initial}_1}, ..., P_{\text{initial}_n}\} \).

After generating the initial plan, the \textit{monitoring} mechanism supervises the process and relays signals to the \textit{reviewing} mechanism for evaluation and validation.
The reviewing mechanism evaluates the plan through two key checks: 
First, it verifies if the generated content satisfies the task-specific constraints $\mathcal{T}$.
Second, it checks the plan’s structure for any syntax errors and applies necessary corrections.
If any issues are detected, a revision process is triggered to refine the plan:
\begin{align}
    \mathcal{P}_\text{revised}&  \gets \text{PlanRevise}(p_\text{revise}, \mathcal{P}_\text{initial}), \\  
 \mathcal{P} & \gets \text{FormatRevise}(\mathcal{P}_\text{revised}),
\end{align}
where \( p_\text{revised} \) includes the revision prompt for the task instructions \( \mathcal{T} \).
This iterative refinement ensures that the final plan is not only of high quality but also optimally structured to guide robust and effective content generation.

\begin{algorithm}[htb]
\caption{CogWriter Algorithm}
\begin{algorithmic}[1]
\Require Prompts $p_{*}$ including task instruction $\mathcal{T}$
\Ensure Final text $\mathcal{D} = \{D_1, \dots, D_n\}$

\Function{PlanningAgent}{$p_{*}$}
    \State $\mathcal{P}_\text{initial} \gets \text{GenerateInitialPlan}(p_{\text{plan}})$ 
    \State $\mathcal{P}_\text{revised} \gets \text{PlanRevise}(p_{\text{revise}},\mathcal{P}_\text{initial})$
    \State $\mathcal{P} \gets \text{FormatRevise}(\mathcal{P}_\text{revised})$
    \State \Return $\mathcal{P}$
\EndFunction

\Function{GenerationAgents}{$p_{*}, \mathcal{P}$}
    \State Initialize empty document collection $\mathcal{D}$
        \For{each $P_i$ in $\mathcal{P}$}
            \State $P_i' \gets \text{PlanAdjust}(p_{\text{adjust}_i},P_i)$
            \State $D_{\text{initial}_i} \gets \text{Generate}(p_{\text{write}}, P_i')$
            \State $D_i \gets \text{LengthRevise}(p_{\text{length}},D_{\text{initial}_i})$
        \EndFor
        \State $\mathcal{D} \gets \mathcal{D} \cup D_i$
    \State \Return $\mathcal{D}$
\EndFunction

\end{algorithmic}
\end{algorithm}

\subsection{Generation Agents}

Once the global plan \( \mathcal{P} = \{P_{1}, ..., P_{_n}\} \) is finalized by the Planning Agent, multiple Generation Agents take over, each responsible for generating content for a specific description task \( D_i \). 
The process begins with validating and refining the local plan \( P_i \), through \textit{monitoring} and \textit{reviewing} similar to the Planning Agent to ensure it aligns with the instruction requirements \( \mathcal{T} \). 
Concretely, if discrepancies are detected, adjustments are applied to update the plan, as shown in the following equation:  
\begin{equation}
    P_i' \gets \text{PlanAdjust}(p_{\text{adjust}_i},P_i),
\end{equation}  
where $p_{\text{adjust}_i}$ encompasses the specialized prompt designed for reviewing each local plan $P_i$ against the residual informing from $\mathcal{T}$.

Upon validation of \( P_i' \), the agent generates content by executing the plan:  
\begin{equation} 
    D_{\text{initial}_i} \gets \text{Generate}(p_{\text{write}}, P_i'),
\end{equation}   
where \( p_{\text{write}} \) is the prompt to generate content following the guidance of the plan \( \mathcal{P}_i' \). 
Based on our preliminary study, this process generally produces content that meets most instruction criteria. 
However, length constraints may still require further refinement due to the limitations of most current LLMs in controlling output length precisely.
To address this, a revision function adjusts the content to meet the specified length \( L \):  
\begin{equation}
    D_i \gets \text{LengthRevise}(p_{\text{length}},D_{\text{initial}_i}),
\end{equation}  
where \( p_{\text{length}} \) is the prompt used to adjust the content length to \( L \) by expanding or compressing the generated text while preserving key details, semantic integrity, and overall coherence.

By following this process, each segment \(D_i \) seamlessly integrates with the overall narrative structure, ensuring both local coherence and global thematic consistency.

\begin{table*}[t]
  \centering
  \resizebox{\textwidth}{!}{
  \begin{tabular}{lcccccc}
    \toprule
    \textbf{Model} & \textbf{Comp. Rate} & \textbf{Acc. Once} & \textbf{Acc. Range} & \textbf{Acc. Periodic} & \textbf{Avg. Acc.} & \textbf{Words (Req. $\geq$12700)} \\
    \midrule
    LongWriter-Llama3.1-8B       & 0.46         & 0.36         & 0.56         & 0.17         & 0.36         & 11036        \\
    Llama-3.1-8B-Instruct        & 0.94         & 0.36         & 0.49         & 0.17         & 0.34         & 8804         \\
    Llama-3.1-70B-Instruct       & 0.79         & 0.50         & 0.51         & 0.18         & 0.39         & 8055         \\
    Mixtral-8x7B-Instruct-v0.1   & 0.83         & 0.42         & 0.45         & 0.24         & 0.37         & 8113         \\
    Qwen-2-72B-Instruct          & 0.94         & 0.42         & 0.44         & 0.14         & 0.33         & 8013         \\
    \midrule  \midrule
    GPT-4o-mini                  & 0.97         & 0.54         & 0.48         & 0.16         & 0.39         & 8940         \\ 
    + SELF-REFINE                & 0.84         & 0.57         & 0.32         & 0.20         & 0.36         & 8154         \\
    + CoT                         & 0.93         & 0.59         & 0.48         & 0.18         & 0.42         & 10137        \\
    + \textbf{CogWriter (Ours)}                  & \textbf{1.00} (↑0.03) & \textbf{0.74} (↑0.20) & \textbf{0.61} (↑0.13) & \textbf{0.31} (↑0.15) & \textbf{0.55} (↑0.16) & \textbf{12484} (↑3544) \\
    
    \midrule 
    Qwen-2.5-14B-Instruct        & 0.29         & 0.53         & 0.54         & 0.24         & 0.44         & 1817         \\
    + SELF-REFINE                & 0.17         & 0.45         & 0.63         & 0.21         & 0.43         & 1122         \\
    + CoT                         & 0.30         & 0.46         & 0.20         & 0.16         & 0.27         & 1619         \\
    + \textbf{CogWriter (Ours)}                    & \textbf{0.79} (↑0.51) & \textbf{0.70} (↑0.17) & \textbf{0.65} (↑0.11) & \textbf{0.47} (↑0.23) & \textbf{0.61} (↑0.17) & \textbf{10091} (↑8274) \\
    
    \midrule
    Llama-3.3-70B-Instruct       & 0.99         & 0.59         & 0.63         & 0.21         & 0.48         & 9431         \\
    + SELF-REFINE                & 0.93         & 0.59         & 0.64         & 0.28         & 0.50         & 8491         \\
    + CoT                         & 1.00         & 0.62         & 0.62         & 0.21         & 0.48         & 9302         \\
    + \textbf{CogWriter (Ours)}                    & \textbf{1.00} (↑0.01) & \textbf{0.76} (↑0.17) & \textbf{0.79} (↑0.16) & \textbf{0.55} (↑0.34) & \textbf{0.70} (↑0.22) & \textbf{12051} (↑2620) \\
    
    \midrule 
    GPT-4o                       & 0.63         & 0.63         & 0.60         & 0.17         & 0.47         & 9055         \\
    + SELF-REFINE                & 0.66         & 0.67         & 0.62         & 0.33         & 0.54         & 4641         \\
    + CoT                         & 0.40         & 0.58         & 0.63         & 0.32         & 0.51         & 4482         \\
    + \textbf{CogWriter (Ours)}                    & \textbf{0.91} (↑0.29) & \textbf{0.80} (↑0.17) & \textbf{0.76} (↑0.16) & \textbf{0.67} (↑0.50) & \textbf{0.74} (↑0.27) & \textbf{11618} (↑2563) \\
    \bottomrule
  \end{tabular}%
  }
  \caption{Model Performance Comparison and the Improvement Brought by CogWriter (values in parentheses indicate the improvement relative to the base model).}
  \label{performance}
\end{table*}

\section{Experiments}
\subsection{Experimental Setup}

\paragraph{Dataset} We evaluated CogWriter using Long- GenBench-16K~\cite{wu2024longgenbench}, a benchmark specifically designed for assessing a model's complex constrained long-form text generation capabilities. 
The dataset features four scenarios, each requiring approximately 16,000 tokens: (1) Diary Writing and (2) Menu Design assess temporal consistency by requiring coherent content organization across weeks of a year, while (3) Skyscraper Design and (4) Urban Planning evaluate spatial reasoning through detailed facility arrangements across floors or city blocks. 
The benchmark includes 400 test instances, with 100 instances per scenario.
Each scenario involves three instruction types (defined in Section~\ref{probdef}): single instructions, range instructions, and periodic instructions. For temporal tasks, Diary Writing and Menu Design require at least 200 words per weekly entry, totaling 10,400 words (52 weeks × 200 words). For spatial tasks, Skyscraper Design and Urban Planning mandate 15,000 words (100 units × 150 words).

\paragraph{Evaluation Metrics} We evaluate model performance using three key metrics from LongGenBench. \textit{Main Task Completion Rate} (Comp. Rate) assesses whether all designated subtasks are completed in sequence (e.g., generating entries for every week in a diary without omissions). \textit{Instruction Following Accuracy} measures adherence to single (Acc. Once), range (Acc. Range), and periodic (Acc. Periodic) instructions, with their average reported as Avg. Acc. We utilize the official evaluation scripts to ensure consistency with reported benchmarks. Additionally, we track \textit{Word Count}, ensuring a minimum average threshold of 12,700 words to meet the combined task requirements.

\paragraph{Experimental Setup} We evaluate our approach across three categories of models and methods. First, we establish baseline performance using several single-pass generation models from the official LongGenBench repository, including LongWriter-Llama3.1-8B~\cite{bai2024longwriterunleashing10000word}, Llama-3.1-8B-Instruct, Mixtral-8x7B-Instruct-v0.1~\cite{jiang2023mistral7b}, Llama-3.1-70B~\cite{grattafiori2024llama3herdmodels}, Qwen-2-72B-Instruct~\cite{qwen2025qwen25technicalreport}, as well as GPT-4o and GPT-4o-mini.
Second, we compare against two prominent enhancement methods: SELF-REFINE~\cite{madaan2023selfrefine} and Chain-of-Thought (CoT) prompting~\cite{wei2022chain}. These methods are applied to four representative foundation models to ensure comprehensive evaluation across different model capabilities and architectures.
Finally, to demonstrate the effectiveness of our CogWriter paradigm, we apply it to the same four foundation models: GPT-4o-mini-2024-07-18, GPT-4o-2024-08-06, Qwen-2.5-14B~\cite{qwen2.5}, and Llama-3.3-70B~\cite{touvron2024llama}. This selection encompasses closed-source and open-source models with varying parameter scales, enabling us to evaluate CogWriter's generalizability. For fair comparison, we implement SELF-REFINE and CoT baselines on these same models alongside our proposed framework.

\begin{table*}[ht!]
  \centering
  \label{tab:performance}
  \resizebox{\textwidth}{!}{\begin{tabular}{lcccccc}
    \toprule
    \textbf{Model} & \textbf{Comp. Rate} & \textbf{Acc. Once} & \textbf{Acc. Range} & \textbf{Acc. Periodic} & \textbf{Avg. Acc.} & \textbf{Words (Req. $\geq$12700)} \\
    \midrule
    GPT-4o-mini + CogWriter & \textbf{1.00} & \textbf{0.74} & \textbf{0.61} & 0.31 & \textbf{0.55} & \textbf{12484} \\
    \midrule
    - w/o PlanRevise  &   0.99       &  0.73        & 0.45         &   0.33     &   0.50       &    12472     \\
    - w/o PlanAdjust  &  \textbf{1.00}       &  0.63        & 0.46         &   0.27      &   0.45       &    12341     \\
    - w/o LengthReview  &  \textbf{1.00}       &   0.73       &    0.61      &   0.30      &   0.54       &  11549       \\
    \bottomrule
  \end{tabular}}
  \caption{Ablation study on the effectiveness of CogWriter's key components using GPT-4o-mini as the base model.}
  \label{ablation}
\end{table*}

\paragraph{Implementation Details} We deployed our experiments across local computational resources and cloud-based APIs. 
For open-source models (Qwen-2.5-14B and Llama-3.3-70B), we leveraged vLLM~\cite{kwon2023efficient} for its efficient inference acceleration while maintaining the default temperature and sampling parameters as specified in the official Hugging Face implementations. 
These experiments were conducted on 4 NVIDIA A100-SXM4-80GB GPUs running CUDA 12.8. 
For closed-source models (GPT-4o and GPT-4o-mini), we utilized their respective official API.

\subsection{Main Results}
Table~\ref{performance} highlights the main performance outcomes of our experiments. 
Firstly, our results reveal that \textit{LongWriter-Llama3.1-8B, despite being specifically designed and trained from Llama-3.1-8B-Instruct for long-form generation, struggles considerably}, achieving only a 0.46 completion rate. 
Similarly, even advanced models with substantial parameter counts, such as Llama-3.1-70B-Instruct and Qwen-2-72B-Instruct, fail to reach the target length of 12,700 tokens in their generated outputs.
Secondly, \textit{alternative enhancement methods also exhibit limited effectiveness.}
Chain-of-Thought prompting results in a modest improvement in instruction-following accuracy (from 0.39 to 0.42 using GPT-4o-mini), while SELF-REFINE achieves reasonable completion rates. 
However, both approaches fall short in meeting length requirements and maintaining instruction adherence.

In contrast, CogWriter demonstrates remarkable improvements across all evaluation metrics. 
When using Qwen-2.5-14B-Instruct as its backbone, it boosts the completion rate by 0.51 and improves average accuracy by 0.17. For Llama-3.3-70B-Instruct and GPT-4o, CogWriter achieves near-perfect completion rates while consistently enhancing instruction-following accuracy, excelling at handling complex periodic instructions.

\begin{table}[htb] 
\centering
\small

\begin{tabular}{lcccc}
\toprule
\textbf{Method} & \textbf{Plan} & \textbf{Decomp.} & \textbf{Monit.} & \textbf{Rev.} \\
\midrule
Human Writer   & \checkmark & \checkmark & \checkmark & \checkmark \\
CoT            & \checkmark & $\times$ & $\times$ & $\times$ \\
SELF-REFINE    & $\times$ & $\times$ & $\times$ & \checkmark \\
Single-pass LLMs    & $\times$ & $\times$ & $\times$ & $\times$ \\
CogWriter      & \checkmark & \checkmark & \checkmark & \checkmark \\
\bottomrule
\end{tabular}%

\caption{Comparison of different writing approaches. \textit{Plan}: planning the writing structure; \textit{Decomp.}: decomposing complex tasks into manageable components; \textit{Monit.}: monitoring progress during generation; \textit{Rev.}: reviewing and refining generated content.}
\label{tab:comparison}
\end{table}

\paragraph{Advantages of Cognitive Structure}
We provide a comparison of the cognitive capabilities of the baselines, our proposed paradigm, and human writers in Table~\ref{tab:comparison}, to analyze the strong performance of our approach. 
It can be seen that human writers naturally employ all four cognitive processes—planning, decomposition, monitoring, and reviewing—while existing computational methods implement only subsets of these capabilities. 
CoT primarily focuses on planning, and SELF-REFINE incorporates only reviewing.
In contrast, CogWriter mirrors the complete human writing process by integrating all four capabilities, which may help explain its superior effectiveness in complex long-form generation tasks.

\paragraph{Correlation with Model Internal Ability}  
We next discuss the relationship between performance improvements and the model's capabilities. When applying our framework to Llama-3.1-8B-Instruct, we observed a clear limitation: the model struggled to generate coherent and structured plans essential for CogWriter's method. In contrast, for stronger LLMs such as GPT-4o, CogWriter achieved significant improvements, including a 0.29 increase in completion rate and a 0.50 increase in periodic instruction accuracy. 
This suggests that models with more advanced internal cognitive abilities are better at utilizing CogWriter’s coordination of cognitive processes, while weaker models, lacking robust instruction-following skills, fail to fully replicate this process.
This limitation shows that CogWriter’s effectiveness depends on the model’s internal abilities, with advancing LLMs enabling more human-like reasoning and problem-solving.

\section{Discussion}

\paragraph{Ablation Study} 
We conduct an ablation study to evaluate the impact of different components in our proposed CogWriter framework, as shown in Table~\ref{ablation}.  
Removing the \textit{PlanRevise} module resulted in a noticeable performance drop across key metrics, with the average accuracy decreasing from 0.55 to 0.50. 
This demonstrates that refining the initial plan through iterative revisions is crucial for maintaining effective task decomposition and alignment with task-specific constraints.  
Disabling the \textit{PlanAdjust} mechanism further impacted performance, reducing the average accuracy to 0.45, particularly affecting Acc. Once and Acc. Range.  
Finally, removing the \textit{LengthReview} module led to a drop in content generation quality due to unmet length constraints, highlighting its role in fine-tuning the output to meet requirements.  
Overall, the results emphasize the importance of each component, with \textit{PlanRevise} and \textit{PlanAdjust} playing key roles in ensuring task decomposition, plan refinement, and overall accuracy of generation.

\begin{figure}[t!]
\centering
  \includegraphics[width=0.8\linewidth]{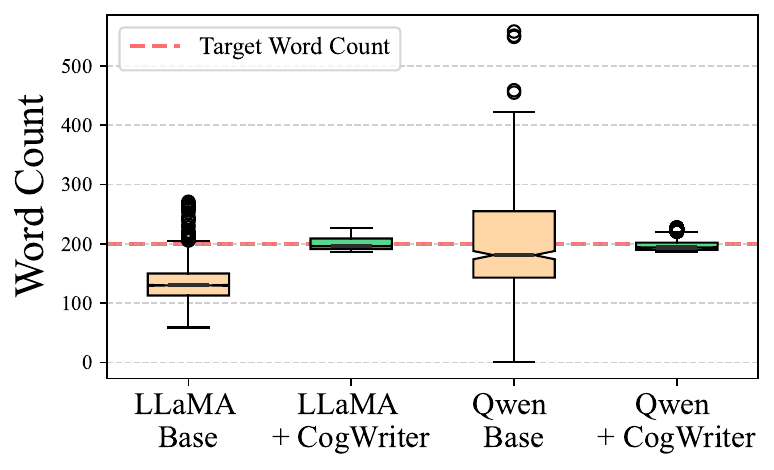} 
  \caption {Comparison of Length Control Ability.}
  \label{compl}
\end{figure}

\paragraph{Length Control Performance}  
As specified in Section~\ref{probdef}, each description \( D_i \) must achieve a target word count of \( L \). 
To evaluate compliance with this requirement, we conducted an analysis of word count distributions across different models. 
Taking the Diary Writing task as an example, Figure~\ref{compl} illustrates the performance of LLama-3.3-70B-Instruct and Qwen-2.5-14B-Instruct. 
The box plot reveals that these base models struggle to meet the word count requirement, with high variance and frequent deviations from the target length.
In contrast, CogWriter achieves superior length control, as shown by its tighter, more stable distribution of word counts.
The explicit monitoring mechanism within CogWriter effectively reduces variance and ensures consistent compliance with the length requirement. 
We provide further analysis results of other models and tasks in Appendix~\ref{lencon}.

\begin{figure}[t!]
\centering
  \includegraphics[width=0.99\linewidth]{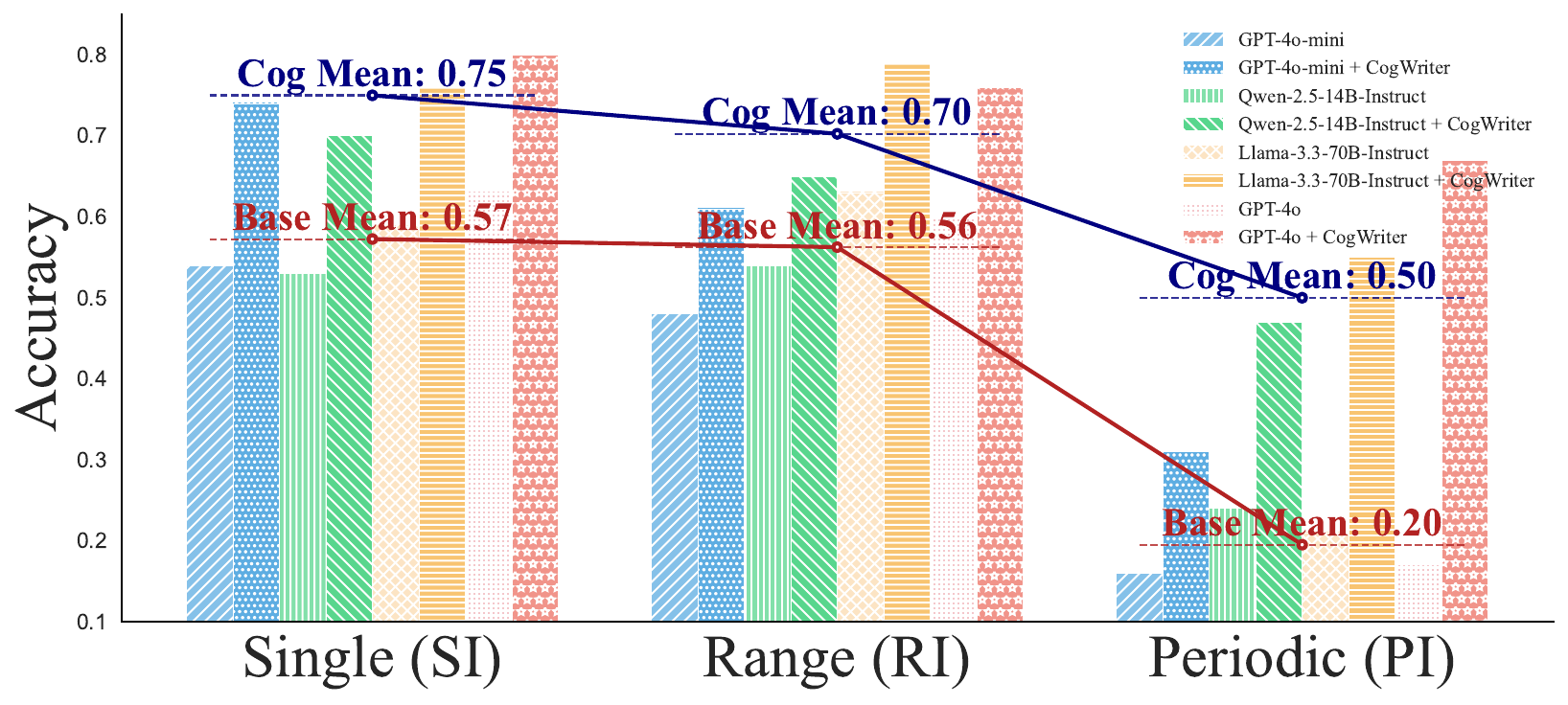} 
  \caption {Comparison of Instruction Type Performance.}
  \label{compt}
\end{figure}

\paragraph{Challenges in Handling Complex Instructions}
As shown in Figure~\ref{compt}, our experiments reveal that for all baselines and our model, the average performance follows a consistent ranking: Single Instructions (SI) outperform Range Instructions (RI), while Periodic Instructions (PI) show the lowest success rate. 
This indicates that, despite task decomposition simplifying the overall process, LLMs still face difficulties in understanding and executing complex instructions. 
One major issue is instruction overload—as the number of instructions increases, the model’s accuracy drops due to the difficulty in managing multiple constraints simultaneously. 
Additionally, instruction complexity plays a significant role: Single Instructions are easier as they target fixed positions, Range Instructions involve more positional flexibility, and Periodic Instructions require tracking repetitions across intervals, making them the most challenging to execute correctly. 
To improve performance in real-world application, it is advisable to limit the number of instructions and manually simplify complex or overlapping instructions where possible.

\section{Related Work}

\paragraph{Long-form Text Generation}
Recent advances in long-form generation have focused on improving models through architectural enhancements and specialized training techniques~\cite{salemi2025experteffectiveexplainableevaluation,que2024hellobenchevaluatinglongtext,liu-etal-2023-task,Li2023TeachLT}. Approaches like Re3~\cite{yang-etal-2022-re3} use recursive reprompting for extended story generation, while DOC~\cite{yang-etal-2023-doc} and hierarchical outlining~\cite{wang2024generatinglongformstoryusing} improve narrative coherence through structured task decomposition. Personalized long-form generation has also gained attention~\cite{salemi2025experteffectiveexplainableevaluation,wang-etal-2024-learning-personalized}, with methods like LongLaMP~\cite{kumar2024longlampbenchmarkpersonalizedlongform} and reasoning-enhanced techniques~\cite{salemi2025reasoningenhancedselftraininglongformpersonalized} adapting models to meet user-specific needs. Similarly, long-form question answering focuses on producing detailed responses to complex queries~\cite{dasigi-etal-2021-dataset,stelmakh-etal-2022-asqa,pmlr-v202-lee23n,tan-etal-2024-proxyqa}. While these methods have improved generation capabilities~\cite{wu2024longgenbench,que2024hellobenchevaluatinglongtext}, our work addresses a critical gap by examining long-form generation through the lens of cognitive writing theory.

\paragraph{Multi-agent Writing}
Multi-agent writing has made notable progress in recent years~\cite{guo2024large,liu2024skepticism,song2024hazards}, showing how agents can collaborate on diverse writing tasks~\cite{10.1007/s11704-024-40231-1,hong2024metagpt}. Research has explored heterogeneous agent integration~\cite{chen2025internet} and educational applications~\cite{10410857}. In academic writing, frameworks like SciAgents~\cite{Ghafarollahi2024SciAgentsAS} demonstrate collaboration among specialized agents for complex writing tasks~\cite{2024autosurvey,DArcy2024MARGMR,su2024headsbetteronemultiagent}, while the Agents' Room approach~\cite{huot2024agentsroomnarrativegeneration} highlights the value of task decomposition in narrative writing.
Beyond academic contexts, multi-agent methods have been applied to creative and informational writing, such as Wikipedia-style articles~\cite{shao-etal-2024-assisting} and poetry~\cite{zhang2024llmbasedmultiagentpoetrygeneration,chen-etal-2024-evaluating-diversity}. 
While these methods focus on collaboration, our work applies cognitive writing principles with agents for planning, monitoring, and revisions, enabling flexible adaptation without task-specific training.

\section{Conclusion and Future Work}

In this paper, we analyzed the challenges of constrained long-form text generation from a cognitive writing perspective. Building on these insights and empirical observations, we proposed CogWriter, a novel writing framework that transforms LLM constrained long-form text generation into a systematic cognitive paradigm. CogWriter bridges the gap between human writing cognition and LLM capabilities, leading to substantial and consistent improvements in both instruction completion and generation length across different LLMs, as demonstrated through extensive experiments on LongGenBench. Looking forward, we plan to optimize agent communication cost and develop specialized models that better align with the unique requirements of each cognitive stage in the writing process.


\section*{Limitations}
While demonstrating superior performance, CogWriter exhibits two primary limitations. First, while our approach achieves higher quality output, it necessitates more computational resources. As detailed in Appendix~\ref{sec:efficiency}, this additional cost stems from multiple rounds of planning, generation, and reviewing. Second, our current implementation utilizes a single LLM across all cognitive writing stages (planning, generation, and reviewing). This uniform approach may not fully leverage the model's capabilities, as each stage only activates specific aspects of the model's knowledge and abilities. Future research directions include exploring specialized models for different cognitive stages and investigating Mixture-of-Experts architectures to enhance both domain expertise and parameter efficiency in the cognitive writing process.

\section*{Ethical Considerations}
Like other LLMs, our CogWriter framework may inherit biases from training data. It may generate inaccurate content despite its enhanced control mechanisms, emphasizing the need for human oversight in practical applications. While the multi-step cognitive process increases computational costs, the structured planning approach improves efficiency and could be further optimized for sustainability. As with any advanced text generation system, CogWriter could potentially be misused for generating deceptive content, highlighting the importance of responsible deployment and appropriate safeguards in real-world applications.

\bibliography{custom}

\begin{figure*}[!t]
    \centering
    \begin{subfigure}[b]{0.45\textwidth}
        \centering
        \includegraphics[width=\textwidth]{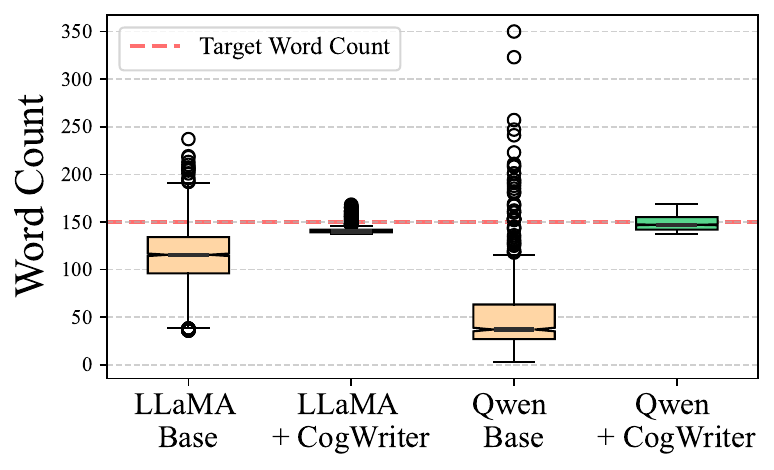}
        \caption{Llama and Qwen on Spatial Tasks}
        \label{fig:length_150_lq}
    \end{subfigure}
    \hfill
    \begin{subfigure}[b]{0.45\textwidth}
        \centering
        \includegraphics[width=\textwidth]{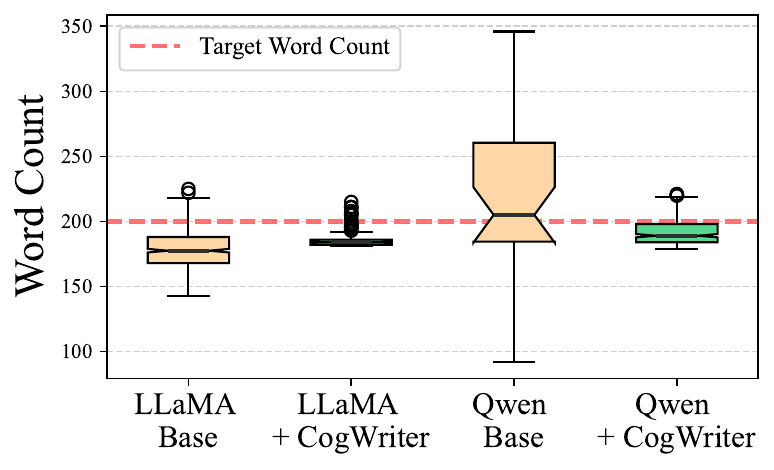}
        \caption{Llama and Qwen on Temporal Tasks}
        \label{fig:length_200_lq}
    \end{subfigure}
    \vskip\baselineskip
    \begin{subfigure}[b]{0.45\textwidth}
        \centering
        \includegraphics[width=\textwidth]{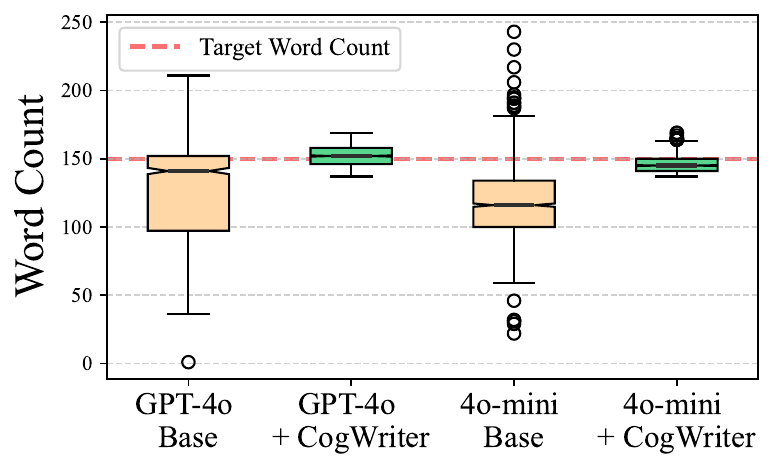}
        \caption{GPT-4o and GPT-4o-mini on Spatial Tasks}
        \label{fig:length_150_gpt}
    \end{subfigure}
    \hfill
    \begin{subfigure}[b]{0.45\textwidth}
        \centering
        \includegraphics[width=\textwidth]{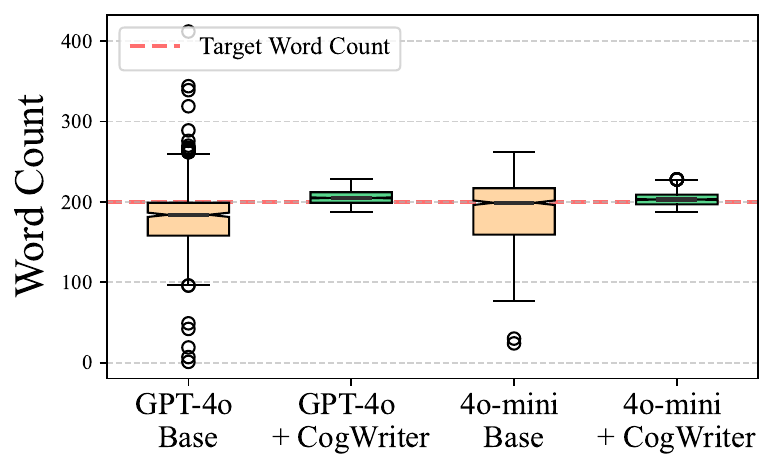}
        \caption{GPT-4o and GPT-4o-mini on Temporal Tasks}
        \label{fig:length_200_gpt}
    \end{subfigure}
    \caption{Length Control Performance Across Different Models and Task Types. (a) and (c) show performance on spatial tasks requiring 150 words per unit, while (b) and (d) present results for temporal tasks with 200-word requirements.}
    \label{fig:length_control}
\end{figure*}

\appendix
\section{Appendix}
\subsection{Further Length Control Performance}
\label{lencon}


To comprehensively demonstrate CogWriter's length control capabilities across different scenarios, we present the generated length distribution of LLama-3.3-70B-Instruct, Qwen-2.5-14B-Instruct, GPT-4o, and GPT-4o-mini in Figures~\ref{fig:length_150_lq}-\ref{fig:length_200_gpt}. 
We evaluate two distinct task types: spatial tasks (150 words) and temporal tasks (200 words). 
Spatial tasks, such as Skyscraper Design and Urban Planning, require detailed facility arrangements across floors or city blocks, with a target length of 150 words per unit. 
In contrast, temporal tasks, including Diary Writing and Menu Design, emphasize temporal consistency across weeks of a year and require 200 words per weekly entry.
Figures~\ref{fig:length_150_lq} and~\ref{fig:length_150_gpt} illustrate model performance on spatial tasks, while Figures~\ref{fig:length_200_lq} and~\ref{fig:length_200_gpt} present results for temporal tasks, highlighting the models' ability to adhere to different length constraints across varying task structures.


\subsection{Inference Time and Token Consumption Analysis}
\label{sec:efficiency}
To evaluate and analyze the computational efficiency of CogWriter, we conducted comprehensive experiments examining inference time and token consumption amount. 

\begin{figure}[ht!
]
\centering
\includegraphics[width=0.9\linewidth]{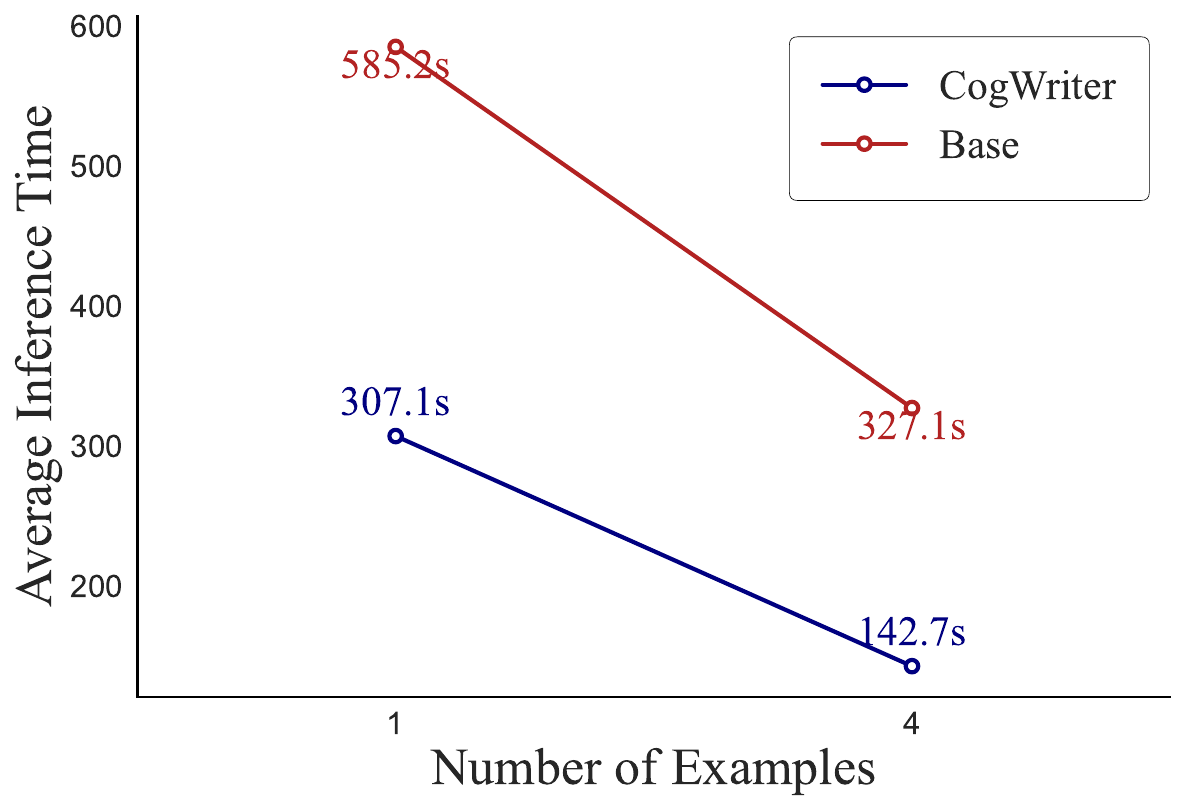}
\caption{Inference Time Comparison.}
\label{fig:inference_time}
\end{figure}

\textbf{Inference Time} For ensure reliable evaluation, we used LLaMA-3.3-70B as our test model, as Qwen exhibited incomplete text generation issues and GPT's API calls were subject to network latency variations. All experiments were performed on 4 NVIDIA A100 GPUs, with each condition tested three times to ensure reliable results.
The experiments were structured as follows: 1) Single text condition: One randomly sampled writing task and 2) 4-example condition: One randomly sampled example from each of the four tasks.
We leveraged vLLM for inference acceleration while maintaining default temperature and sampling parameters from official Hugging Face implementations. To ensure a fair comparison, we only considered outputs achieving 100\% completion rate.
Figure~\ref{fig:inference_time} illustrates the inference time comparison between CogWriter and the baseline model across different batch sizes.
 
Through the implementation of multi-generation agents for parallel processing, our approach demonstrates a significant reduction in generation time, achieving approximately 50\% faster processing compared to the baseline model.

\textbf{Token Consumption} Our analysis reveals that CogWriter consumes approximately 2.8 times more output tokens and 10 times more total tokens compared to baseline methods. The observed increase in token utilization can be attributed to two primary factors:

\begin{enumerate}
    \item While CogWriter ensures comprehensive output generation, baseline models frequently produce responses that are incomplete in quality and length. Notably, baseline models such as GPT-4o often acknowledge their limitations with responses like ``\textit{I'm sorry, but creating an entire year's worth of weekly diary entries with detailed narratives is beyond my capabilities in a single response},'' resulting in artificially lower token consumption metrics.
    
    \item CogWriter employs an iterative approach involving multiple rounds of plan evaluation against the original prompt, analogous to the human writing process where additional cognitive effort correlates with enhanced document quality and comprehensiveness, thereby increasing token usage.
\end{enumerate}

Despite these considerations, it is noteworthy that while GPT-4o's API pricing is 16.67 times higher than GPT-4o-mini\footnote{\url{https://openai.com/api/pricing/}}, it achieves only a marginal improvement in Average Accuracy (0.08), as demonstrated in Table~\ref{performance}. In contrast, CogWriter demonstrates a more substantial improvement of 0.16 in Average Accuracy over GPT-4o-mini. Furthermore, our framework can be implemented with lightweight closed-source models such as Qwen-2.5-14B-Instruct, enabling local deployment. This capability is particularly valuable for applications prioritizing output quality and data privacy, including professional content creation, academic writing, and technical documentation.

Our research primarily focuses on transcending the limitations inherent in conventional single-pass generation approaches, aiming to achieve text quality that surpasses the capabilities of individual LLMs, including advanced models like GPT-4o. Much like professional writing practices, where quality content necessitates extended development time and thinking compared to preliminary drafts, CogWriter's increased resource utilization reflects the sophistication of its cognitive processing mechanisms.

While acknowledging the additional computational overhead, we identify several promising directions for future research, including the development of memory optimization techniques and the exploration of specialized writing models with enhanced parameter efficiency for specific cognitive processes in the generation pipeline.

\end{document}